# Predicting drug response of tumors from integrated genomic profiles by deep neural networks

Yu-Chiao Chiu[1], Hung-I Harry Chen[1,2], Tinghe Zhang[2], Songyao Zhang[2,3], Aparna Gorthi[1], Li-Ju Wang[1], Yufei Huang[2,4§], Yidong Chen[1,4§]

[1]Greehey Children's Cancer Research Institute, University of Texas Health Science Center at San Antonio, San Antonio, TX 78229, USA

[2]Department of Electrical and Computer Engineering, University of Texas at San Antonio, San Antonio, TX 78249, USA

[3]Laboratory of Information Fusion Technology of Ministry of Education, School of Automation, Northwestern Polytechnical University, Xi'an, Shaanxi 710072, China

[4]Department of Epidemiology and Biostatistics, University of Texas Health Science Center at San Antonio, San Antonio, TX 78229, USA

[§]Corresponding authors

Email addresses:

    YH: Yufei.Huang@utsa.edu

    YC: ChenY8@uthscsa.edu

# Abstract

**Background**

The study of high-throughput genomic profiles from a pharmacogenomics viewpoint has provided unprecedented insights into the oncogenic features modulating drug response. A recent study screened for the response of a thousand human cancer cell lines to a wide collection of anti-cancer drugs and illuminated the link between cellular genotypes and vulnerability. However, due to essential differences between cell lines and tumors, to date the translation into predicting drug response in tumors remains challenging. Recently, advances in deep neural networks (DNNs) have revolutionized bioinformatics and introduced new techniques to the integration of genomic data. Its application on pharmacogenomics may fill the gap between genomics and drug response and improve the prediction of drug response in tumors.

**Results**

We proposed a DNN model to predict drug response based on mutation and expression profiles of a cancer cell or a tumor. The model contains three subnetworks, i) a mutation encoder pre-trained using a large pan-cancer dataset to abstract core representations of high-dimension mutation data, ii) a pre-trained expression encoder, and iii) a drug response predictor network integrating the first two subnetworks. Given a pair of mutation and expression profiles, the model predicts $IC_{50}$ values of 265 drugs. We trained and tested the model on a dataset of 622 cancer cell lines and achieved an overall prediction performance of mean squared error at 1.96 (log-scale $IC_{50}$ values). The performance was superior in prediction error or stability than two classical methods (linear regression and support vector machine) and four analog DNNs of our model, including DNNs built without TCGA pre-training, partly replaced by principal components, and built on individual types of input data. We then applied the model to predict drug response of 9,059 tumors of 33 cancer types. Using per-cancer and pan-cancer settings, the model predicted both known, including EGFR inhibitors in non-small cell lung cancer and tamoxifen in ER+ breast cancer, and novel drug targets, such as vinorelbine for *TTN*-mutated tumors. The comprehensive analysis further revealed the molecular mechanisms underlying the resistance to a chemotherapeutic drug docetaxel in a pan-cancer setting and the anti-cancer potential of a novel agent, CX-5461, in treating gliomas and hematopoietic malignancies.

**Conclusions**

Here we present, as far as we know, the first DNN model to translate pharmacogenomics features identified from *in vitro* drug screening to predict the response of tumors. The results covered both well-studied and novel mechanisms of drug resistance and drug targets. Our model and findings improve the prediction of drug response and the identification of novel therapeutic options.

## Background

Due to tumor heterogeneity and intra-tumor sub-clones, an accurate prediction of drug response and an identification of novel anti-cancer drugs remain challenging tasks [1, 2]. Pharmacogenomics, an emerging field studying how genomic alterations and transcriptomic programming determine drug response, represents a potential solution [3, 4]. For instance, recent reports identified mutation profiles associated with drug response both in tumor type-specific and pan-cancer manners [5, 6]. As drug response data of large patient cohorts are scarcely available, large-scale cell line-based screening can greatly facilitate the study of pharmacogenomics in cancer. Recently, the Genomics of Drug Sensitivity in Cancer (GDSC) Project proposed a comprehensively landscape of drug response of ~1,000 human cancer cell lines to 265 anti-cancer drugs and unveiled crucial oncogenic aberrations related to drug sensitivity [7, 8]. Because of the fundamental differences between *in vitro* and *in vivo* biological systems, a translation of pharmacogenomics features derived from cells to the prediction of drug response of tumors is to our knowledge not yet realized.

Deep learning (DL) is the state-of-the-art machine learning technology for learning knowledge from complex data and making accurate predictions. It features the ability to learn the representation of data without the need for prior knowledge and an assumption on data distributions. The DL technology has been successfully applied to bioinformatics studies of regulatory genomics, such as predicting binding motifs [9], investigating DNA variants [10], deciphering single-cell omics [11, 12], and extraction of genomics features for survival prediction [13]. In pharmaceutical and pharmacogenomics research, reports have shown its ability to predict drug-target interactions [14], screen for novel anti-cancer drugs [15], and predict drug synergy [16]. Nevertheless, data complexity and the requirement of large training datasets have limited its application to integrate genomics data and comprehensively predict drug response, hindering the translation to precision oncology.

Addressing the unmet demands, the present study is aimed to predict the response of tumors to anti-cancer drugs based on genomic profiles. We designed a deep neural network (DNN) model to learn the genetic background from high-dimensional mutation and expression profiles using the huge collection of tumors of The Cancer Genome Atlas (TCGA). The model was further trained by the pharmacogenomics data developed in human cancer cell lines by the GDSC Project and their corresponding genomic and transcriptomic alteration, and finally applied to TCGA data again to predict drug response of tumors. Collectively, this study demonstrated a novel DL model that bridges cell line-based pharmacogenomics knowledge via tumor genomic and transcriptomic abstraction to predict tumors' response to compound treatment.

## Methods

**Datasets**

We downloaded gene-level expression data of 935 cell lines of the Cancer Cell Line Encyclopedia (CCLE) and 11,078 TCGA pan-cancer tumors from the CTD[2] Data Portal [17] and UCSC TumorMap [18], respectively. Given the total numbers of cell lines, tumors, and genes as *C*, *T*, *G*, respectively, we metricized the expression data by $E^{CCLE} = \{log_2(tpm_{g,c}^{CCLE} + 1)\}$, where $tpm_{g,c}^{CCLE}$ is the number of transcripts per million of gene *g* ($g \in [1, G]$) in cell line *c* ($c \in [1, C]$), and $E^{TCGA} = \{log_2(tpm_{g,t}^{TCGA} + 1)\}$, where $tpm_{g,t}^{TCGA}$ denotes the number of transcripts per million of the same gene in tumor *t* ($t \in [1, T]$). Genes with low information burden (mean < 1 or st. dev. < 0.5) among TCGA samples were removed. Mutation Annotation Format (MAF) files of mutation data were downloaded directly from CCLE (1,463 cells) [19, 20] and TCGA databases (10,166 tumors). Here we only considered four types of nonsynonymous mutations, including missense and nonsense mutations, frameshift insertions and deletions. Thus, we had binary matrices of $M^{CCLE} = \{m_{g,c}^{CCLE}\}$ and $M^{TCGA} = \{m_{g,t}^{TCGA}\}$, where $m_{g,c}^{CCLE}$ and $m_{g,t}^{TCGA}$ are the mutation states (1 for mutation and 0 for wildtype) of gene *g* in *c* and *t*, respectively. Genes with no mutations in CCLE and TCGA samples were eliminated.

We also downloaded drug response data of 990 CCLE cell lines to 265 anti-cancer drugs measured by the half maximal inhibitory concentration (IC$_{50}$) from the GDSC Project [7]. IC$_{50}$ were measured in μM and represented in log scale (*i.e.*, $IC^{CCLE} = \{log_{10}(ic_{d,c}^{CCLE})\}$, with *d* denoting the *d*-th drug and $d \in [1, D]$) and missing data were imputed by a weighted mean of IC$_{50}$ of 5 nearest drugs using R packages VIM and laeken [21, 22]. In this study, we analyzed 622 cell lines with available expression, mutation, and IC$_{50}$ data and 9,059 tumors with expression and mutation profiles.

**General settings of DNNs and computation environment**

DNN training in this study were performed using the python library Keras 1.2.2 with TensorFlow backend. We used fully (or densely) connected layers for all networks. At a neuron $j$, its output $y_j$ is calculated by

$$y_j = F\left(\sum_i w_{ij} x_i + b_j\right) \qquad (1)$$

, where $x_i$ is the output of neuron $i$ at the previous layer of $j$, $w_{ij}$ and $b_j$ denote the synaptic weight and bias, respectively, and $F$ represents an activation function. The notation of all neurons at a layer can thus be written as

$$\boldsymbol{y} = F(\boldsymbol{wx} + \boldsymbol{b}). \qquad (2)$$

During training, synaptic weights and biases are adjusted to minimize a loss function. We hereafter refer to the two parameters as synaptic parameters because they represent the model and can be used to transfer a learned model to another. In this study, models were optimized using the Adam optimizer with a loss function of mean squared error (MSE). We used the He's uniform distribution [23] to initialize autoencoders and the Prediction (P) network, while the mutation encoder (M$_{enc}$) and expression encoder (E$_{enc}$) in the complete model were initialized by the synaptic parameters learned from the pre-training on TCGA data. Neuron activation function was set as rectified linear unit (ReLU) except for the output layer of P as linear in order to better fit the distribution of log-scale IC$_{50}$.

**Overview of the proposed DNN model**

The proposed DNN model was developed to predict $IC_{50}$ values based on genomic profiles of a cell or a tumor. Given the pair of mutation and expression vectors of sample $c$, $\{\boldsymbol{M^{CCLE}}(:,c), \boldsymbol{E^{CCLE}}(:,c)\}$, the model predicts a $D$-length vector of $IC_{50}$, $\widehat{\boldsymbol{IC^{CCLE}}}(c)$, as an output. As shown in Figure 1, the model is composed of three networks: i) a mutation encoder ($M_{enc}$), ii) an expression encoder ($E_{enc}$), and iii) a prediction feedforward network (P). The first and second components are the encoding parts of two autoencoders pre-trained using TCGA data to learn the high-order features of mutation and expression data into a lower dimensional representation. The encoded representation of mutation and expression profiles were linked into P and the entire model was trained on CCLE data to make prediction of $IC_{50}$ values. Details of our model are described below.

**Pre-training of mutation and expression encoders**

Autoencoder is an unsupervised DL architecture that includes an asymmetric pair of encoder and decoder. By minimizing the loss between input and reconstructed (*i.e.*, decoded) data, it reduces the dimension of complex data and captures crucial features at the bottleneck layer (the layer between encoder and decoder) (Figure 1B, top and bottom panels). We pre-trained an autoencoder on each of the TCGA mutation and expression datasets to optimize the capability to capture high-order features. To determine the optimized architecture, we adopted a hyper-parameter optimization method, namely hyperas [24], to select i) number of neurons at the $1^{st}$ layer (4096, 2048, or 1024), ii) number of neurons at the $2^{nd}$ layer (512, 256, or 128), iii) number of neurons at the $3^{rd}$ layer (the bottleneck layer; 64, 32, or 16), and iv) batch size (128 or 64). Each combination was trained for 20 epochs; the best-performing model was re-run for 100 epochs and the synaptic parameters were saved.

**Complete prediction network**

In our complete model, encoders of the two optimized autoencoders, *i.e.*, $M_{enc}$ and $E_{enc}$, were linked to P to make predictions of $IC_{50}$ (Figure 1). P is a 5-layer feedforward neural network, including the first layer merging output neurons of the two encoders, three fully connected layers, and the last layer of *d* neurons generating $IC_{50}$ values of *d* drugs (Figure 1B, orange box). In the complete model, architecture (number of layers and neurons at each layer) of $M_{enc}$ and $E_{enc}$ were fixed; their synaptic parameters were initialized using the parameters obtained from pre-training in TCGA and updated during the training process. P was randomly initialized. We trained the entire model using CCLE data, with 80%, 10%, and 10% of samples as training, validation, and testing sets, respectively. We note the validation dataset was used to update model parameters but to stop the training process when the loss in validation set had stopped decreasing for 3 consecutive epochs to avoid model overfitting. Performance of the model was evaluated using the testing samples, *i.e.*, $MSE\left(\widehat{IC^{CCLE}}(:,C_{test}), IC^{CCLE}(:,C_{test})\right)$, where $C_{test}$ denotes the test set of cell lines.

We applied the final model to predict drug response of TCGA tumors. For a tumor *t*, $\{M^{TCGA}(:,t), E^{TCGA}(:,t)\}$ was fed into the model and $\widehat{IC^{TCGA}}(:,t)$ was calculated. A high predicted $IC_{50}$ indicates an adverse response of a patient to the corresponding drug.

**Comparison to other model designs**

Performance of the complete neural network model was compared to four different DNN designs. First, to assess the effect of TCGA pre-training on $M_{enc}$ and $E_{enc}$, we randomly initialized both encoders using the He's uniform distribution and calculated MSE of the entire model. Second, dimension reduction of the $M_{enc}$ and $E_{enc}$ networks was replaced by principal component analysis (PCA). Last two models were built without $M_{enc}$ or $E_{enc}$ to study whether they jointly improved the performance. In each iteration, CCLE samples were randomly assigned to training (80%), validation (10%), and testing (10%) and each model was trained and tested. Performance in terms of the number of consumed epochs and MSE in $IC_{50}$ were summarized and compared across the 100 iterations. We also analyzed two classical prediction methods, multivariate linear regression and regularized support vector machine (SVM). For each method, top 64 principle components of mutations and gene expression were merged to predict $IC_{50}$ values of all (using linear regression) or individual drugs (SVM).

## Results and Discussion

**Model construction and evaluation in CCLE**

The study is aimed to predict drug response (measured as log-scale $IC_{50}$ values) using genome-wide mutation and expression profiles. We included mutation and expression profiles of 622 CCLE cell lines of 25 tissue types and 9,059 TCGA tumors of 33 cancer types. After data preprocessing, 18,281 and 15,363 genes with mutation and expression data, respectively, available in both CCLE and TCGA samples were analyzed. Log-scale $IC_{50}$ values of all cell lines in response to 265 anti-cancer drugs were collected from the GDSC Project [7]. After imputation of missing values, the range of log $IC_{50}$ was from -9.8 to 12.8 with a standard deviation of 2.6 (Figure 2A). We designed a DNN model with three building blocks: 4-layer $M_{enc}$ and 4-layer $E_{enc}$ for capturing high-order features and reducing dimensions of mutation and expression data, and a 5-layer prediction network P integrating the mutational and transcriptomic features to predict $IC_{50}$ of multiple drugs (Figure 1). To make the best use of the large collection of TCGA pan-cancer data, we pre-trained an autoencoder for each data type and extracted the encoders, $M_{enc}$ (number of neurons at each layer, 18,281, 1,024, 256, and 64) and $E_{enc}$ (15,363, 1,024, 256, and 64), to construct our final model (detailed in Methods). Output neurons of the two encoders were linked to P (number of neurons at each layer, 64+64, 128, 128, 128, and 265), of which the last layer outputs predicted $IC_{50}$. Architecture of the complete neural networks is shown in Figure 1B.

After pre-training $M_{enc}$ and $E_{enc}$ components, we trained the entire model using 80% of CCLE samples together with a validation set of 10% of samples to avoid overfitting. The remaining samples (64 cells; 16,960 cell-drug combinations) were used for testing. The model achieved an overall MSE in $IC_{50}$ of 1.53, corresponding to 1.48 and 1.98 in training/validation and testing data, respectively. Generally, the distribution of predicted $IC_{50}$ was similar to original data (Figure 2A-B), while the two modes of original data seemed to be enhanced (highlighted in Figure 2A). In both training/validation and testing data, the prediction was highly consistent to the true data in terms of $IC_{50}$ values (Pearson correlation; $\rho_P$) and rank of drugs (Spearman correlation; $\rho_S$) of a sample ($\rho_P \in [0.70, 0.96]$, $\rho_S \in [0.62, 0.95]$, and all $P$-values $< 1.0 \times 10^{-29}$; Figure 2C-D). Of note, correlations achieved in training/validation and testing samples were highly comparable (Figure 2C-D), confirming the performance of our model.

**Performance comparisons to other designs**

To test the stability of our model, we ran 100 training processes each of which training, validation, and testing cells were reselected. Overall, the model converged in 14.0 epochs (st. dev., 3.5; Table 1) and achieved an MSE of 1.96 in testing samples (st. dev., 0.13; Figure 2E and Table 1). We compared the performance to linear regression, SVM, and four analog DNNs of our model, including random initialization (identical architecture, but without TCGA pre-training of $M_{enc}$ and $E_{enc}$), PCA ($M_{enc}$ and $E_{enc}$ each replaced by top 64 principal components of mutation and expression data), $M_{enc}$ only ($E_{enc}$ removed from the model), and $E_{enc}$ only ($M_{enc}$ removed from the model). The two classical methods seemed to suffer from high MSE in testing samples (10.24 and 8.92 for linear regression and SVM, respectively; Table 1). Our model also outperformed DNNs with random initialization and PCA in MSE (difference in medians, 0.34 and 0.48; Figure 2E and Table 1) and stability (st. dev. of MSE in testing samples = 0.13, 1.21, and 0.17 for our model, random initialization, and PCA, respectively; Figure 2E). While the $E_{enc}$-only model achieved similar performance to our model (difference in medians = 0.0042; Figure 2E and Table 1), the addition of $M_{enc}$ seemed to bring faster convergence (difference in medians = 3; Table 1). Our data echoed the biological premise that gene expressions are more directly linked to biological functions and thus richer in information burden than mutations.

**Associations of gene mutations to predicted drug response in TCGA – per-cancer study**

In search of effective anti-cancer drugs in tumors, we applied the constructed model directly to predict the response of 9,059 TCGA samples to the 265 anti-cancer drugs. The predicted $IC_{50}$ values followed a similar distribution to CCLE cells (Figure 2A, blue line). Realizing the different nature of cell lines and tumors, we started by examining several drugs with well-known target genes. As shown in Figure 3A, breast invasive carcinoma (BRCA) with positive estrogen receptor (ER; assessed by immunohistochemistry by TCGA) responded to a selective estrogen receptor modulator, tamoxifen, significantly better than ER-negative patients (*t*-test $P = 2.3 \times 10^{-4}$). Also, two EGFR inhibitors, afatinib and gefitinib, achieved better performance in non-small cell lung cancers (NSCLC) with mutated *EGFR* ($P = 2.0 \times 10^{-7}$ and $6.6 \times 10^{-3}$). While the promising results on these well-characterized drugs showed the applicability of our model to tumors, we noted that the magnitude of differences in predicted $IC_{50}$ levels was modest, underlining the fundamental differences between cell lines and tumors. In order to prioritize mutations underlying drug response, we systematically analyzed all cancer–mutation–drug combinations and tested the significance of differences in $IC_{50}$ between samples with and without a mutation for each cancer. Here only genes with a mutation rate higher than 10% and harbored by at least 10 patients in a cancer were analyzed. With a stringent criterion of Bonferroni-adjusted *t*- test $P < 1.0\times10^{-5}$, we identified a total of 4,453 significant cancer–mutation–drug combinations involving 256 drugs and 169 cancer–mutation combinations (Figure 3B). The top three combinations were *TP53* mutations in lung adenocarcinoma (LUAD; modulating response to 235 drugs), lung squamous cell carcinoma (LUSC; 228 drugs), and stomach adenocarcinoma (STAD; 224 drugs) (Table 2). *TP53* was one of the most frequently mutated and well-studied genes in many cancers. The mutation has been shown to be associated with cancer stem cells and resistance functions and thus regulates drug resistance [25, 26]. For instance, our data indicated its associations with resistance of a PI3Kβ inhibitor, TGX221, in 9 cancers including low-grade glioma (LGG; mean difference in $IC_{50}$ ($\Delta IC_{50}$) = 0.95; $P = 2.2\times10^{-109}$; Figure 3C) and resistance of vinorelbine in BRCA ($\Delta IC_{50}$ = 0.68; $P = 7.4\times10^{-71}$; Figure 3C) and 6 other cancers. We also identified gene mutations that sensitized tumors to a large number of drugs, such as *IDH1* (138 drugs; Table 2). *IDH1* was the most frequently mutated gene in LGG (77.3% in our data; Table 2) and known to regulate cell cycle of glioma cells and enhance the response to chemotherapy [27]. Our finding agreed with the report and showed that *IDH1* mutation dramatically reduced $IC_{50}$ of chemotherapeutic agents, e.g., doxorubicin in LGG ($\Delta IC_{50}$ = -0.85; $P = 3.6\times10^{-71}$; Figure 3C).

**Associations of gene mutations to predicted drug response in TCGA – pan-cancer study**

We also carried out a study to explore how gene mutations affect drug response in a pan-cancer setting. The analysis was focused on 11 genes with mutation rates higher than 10% across all TCGA samples (Table 3). Using an identical criterion, we identified 2,119 significant mutation–drug pairs composed of 256 drugs, among which 1,882 (88.8%) and 237 (11.2%) were more resistant and sensitive in mutated samples, respectively (Figure 4A and Table 3). *TP53* (251 drugs), *CSMD3* (223), and *SYNE1* (218), *TTN* (206), and *RYR2* (199) were the top drug response-modulating genes (Table 3). Among them, *TP53* (9 sensitive and 242 resistant drugs) and *TTN* mutations (44 and 162) were associated with the most numbers of resistant and sensitive drugs, respectively (Table 3). Thus, we further investigated the drug response and their association with status of the 2 genes. Many of the drugs with large *TP53* mutations-modulated changes in $\Delta IC_{50}$ ($|\Delta IC_{50}| \geq 0.7$; Figure 4A-B) were previously studied in different cancer types by *in vitro* models. For instance, wildtype *TP53* is required in the anti-cancer actions of CX-5461 [28, 29] and sorafenib [30] (both $P$ of $\Delta IC_{50}$ ~0 in our data; Figure 4B), sensitizes various cancer cells to bortezomib [31] ($P = 4.4 \times 10^{-308}$; Figure 4B), and enhances phenformin-induced growth inhibition and apoptosis [32] ($P = 2.0 \times 10^{-241}$; Figure 4B). As for previously less explored *TTN* mutations, the longest gene in human genome known to carry large variations, our data indicated that perhaps *TNN* acts as a marker gene of tumors sensitized to chemotherapeutic agents such as vinorelbine ($P$ ~0; Figure 4C) and a potential anti-cancer drug epothilone B ($P = 2.5 \times 10^{-253}$; Figure 4C). Taken together findings from our per- and pan-cancer studies, we have demonstrated the applicability of our model to predict drug response of tumors and ability to unveil novel and well-studied genes modulating drug response in cancer.

**Pharmacogenomics analysis of docetaxel and CX-5461 in TCGA**

To unveil the pharmacogenomics landscape of drugs, a comprehensive study of mutation and expression profiles associated with resistance of a drug in a pan-cancer setting was carried out. Here we took two drugs as demonstrating examples, a widely used chemotherapeutic agent docetaxel and a novel anti-cancer drug CX-5461 currently under investigation in several cancers. For each drug, pan-cancer patients predicted to be very sensitive and resistant (with $IC_{50}$ in bottom and top 1%, n = 91 in each group; Figure 5A, left panel) were compared for cancer type composition, mutation rates, and differential gene expression. Top cancer types of docetaxel-sensitive patients were among esophageal carcinoma (ESCA; 25.3%), cervical and endocervical cancer (CESC; 13.2%), and head and neck squamous cell carcinoma (HNSC; 9.9%) (Figure 5B, left panel), while top resistant patients were mainly liver hepatocellular carcinoma (LIHC; 42.9%), LGG (26.4%), and glioblastoma multiforme (GBM; 12.1%) (Figure 5B, left panel). Top 10 gene with most changed mutation rates between the two groups of patients are listed in Figure 5C. On average, each sensitive tumor harbored 2.7 mutations among these genes, much higher than 0.51 observed in the resistant group (Figure 5C, left panel), implying tumors with higher mutation burdens in crucial genes may be more vulnerable to the treatment. Of note, a great majority of the most significantly differentially expressed genes were upregulated in sensitive patients (Figure 5C, left panel). We performed functional annotation analysis of the top 300 genes in Gene Ontology terms of biological processes and molecular functions using the Database for Annotation, Visualization and Integrated Discovery (DAVID) v6.7 [33, 34]. While we did not observe any cluster of functions related to microtubule, through which docetaxel physically binds to the cell and regulate the cell cycle [35], these drug sensitivity-related genes were indeed predominantly enriched in functions governing the mitotic cell cycle (Table 4). The observation largely reflected the nature of the chemotherapeutic agent to target highly proliferative cells and the dependence of drug response on the ability to pass cell-cycle checkpoints. In addition to docetaxel, we analyzed a novel anti-cancer agent, CX-5461. This inhibitor of ribosomal

RNA synthesis has been shown with anti-cancer properties in cancer cells [36, 37] and is now under phase I/II clinical trial in solid tumors (NCT number, NCT02719977). In hematopoietic malignancies, it was recently shown to outperform standard chemotherapy regimen in treating aggressive acute myeloid leukemia (LAML) [29], and its anti-cancer effects were dependent on wild-type *TP53* [28, 29]. Concordantly, in our data, LAML and lymphoid neoplasm diffuse large B-cell lymphoma (DLBC) jointly accounted for 45.1% (41.8% and 3.3%) of patients predicted be respond extremely well to CX-5461 (Figure 5A-B, right panels). Of note, LGG comprised another 48.4% of the sensitive tumors (Figure 5B, right panel). Nine of the top 10 differentially mutated genes were enriched in the resistant group and leaded by *TP53* mutations (mutation rate, 95.6% in resistant vs. 13.2% in sensitive patients; Figure 5C, right panel), echoing data from our pan-cancer analysis (Figure 4A-B) and previous *in vitro* and *in vivo* investigations [28, 29]. *IDH1* was the only gene preferentially mutated in sensitive tumors and largely marked LGG (mutated in 42 of 44 sensitive LGG; Figure 5C, right panel). DAVID analysis of the top 300 differentially expressed genes highlighted differential mechanisms between solid and non-solid tumors, such as extracellular matrix and cell motion (Table 5). Altogether, the pharmacogenomics analyses revealed well-known resistance mechanisms of docetaxel and shed light on the potential of CX-5461 on hematopoietic malignancies and LGG.

**Limitations and future work**

DNN is unquestionably one of the hottest computational breakthroughs in the era of big data. Although promising results of our and other studies have demonstrated its ability of solving challenging bioinformatic tasks, the method has several fundamental limitations. For instance, due to high representational power and model complexity, the method suffers from overfitting and the requirement of large training data. Addressing this, the present study adopts a training–validation partition of training data to allow early stopping to the training process [38]. Future work may further incorporate dropout and regularization to DNNs. Also, by taking advantage of the transferability of neural networks, we used the huge volume of TGCA data to equip our model the ability of capturing representations of mutation and expression data. Transferring the learned parameters to initialize our model virtually increased the sample size of our training data. Our data from 100 iterations of model training suggest the stability of performance and insensitivity to the selection of training samples. As the availability of more large-scale drug screening data, we expect the proposed model to make even more accurate predictions and unveil subtle pharmacogenomics features. Furthermore, our model may incorporate additional genomic mutation information, such as copy number alterations, into data matrices $M^{TCGA}$ and $M^{CCLE}$, to enrich the complexity of tumor mutation for model training and further reduce the training MSE. Because of the nature of DNNs as black boxes, the interpretability of results is typically limited. In this study, by integrating genomics profiles to the predictions, we systematically investigated how single gene mutations, as well as the interplay between cancer type, mutations, and biological functions, were associated with the predicted drug response. With the advances in DNN, several novel methods were recently proposed to extract features learned by neural networks, such as network-centric approach [39] and decomposition of predicted outputs by backpropagation onto specific input features [40] (reviewed in [41]). Future works may incorporate these methods to provide a landscape of pharmacogenomics and further reveal novel oncogenic genomics profiles.

# Conclusions

This study addresses the need for a translation of pharmacogenomics features identified from pre-clinical cell line models to predict drug response of tumors. We developed a DNN model capable of extracting representative features of mutations and gene expression, and bridging knowledge learned from cancer cell lines and transferring to tumors. We showed the reliability of the model and its superior performance than four different methods. Applying our model to the TCGA collection of tumors, we identified both well-studied and novel resistance mechanisms and drug targets. Overall, the proposed model is widely applicable to incorporate other omics data and to study a wider range of drugs, paving the way to the realization of precision oncology.

# List of Abbreviations

ACC, adrenocortical cancer; BLCA, bladder urothelial carcinoma; BRCA, breast invasive carcinoma; CCLE, Cancer Cell Line Encyclopedia; CESC, cervical and endocervical cancer; CHOL, cholangiocarcinoma; COAD, colon adenocarcinoma; DL, deep learning; DLBC, diffuse large B-cell lymphoma; DNN, deep neural network; $E_{enc}$, expression encoder; ER, estrogen receptor; ESCA, esophageal carcinoma; GBM, glioblastoma multiforme; HNSC, head and neck squamous cell carcinoma; $IC_{50}$, half maximal inhibitory concentration; KICH, kidney chromophobe; KIRC, kidney clear cell carcinoma; KIRP, kidney papillary cell carcinoma; LAML, acute myeloid leukemia; LGG, lower grade glioma; LIHC, liver hepatocellular carcinoma; LUAD, lung adenocarcinoma; LUSC, lung squamous cell carcinoma; MESO, mesothelioma; $M_{enc}$, mutation encoder; MSE, mean squared error; MUT, mutated; NSCLC, non-small cell lung cancer; Num, number; OV, ovarian serous cystadenocarcinoma; P, prediction network; *P*, *P*-value; PCA, principal component analysis; PCPG, pheochromocytoma and paraganglioma; PRAD, prostate adenocarcinoma; Rand Init, random initialization; READ, rectum adenocarcinoma; SARC, sarcoma; SKCM, skin cutaneous melanoma; STAD, stomach adenocarcinoma; SVM, support vector machine; TCGA, The Cancer Genome Atlas; TGCT, testicular germ cell tumor; THCA, thyroid carcinoma; THYM, thymoma; UCEC, uterine corpus endometrioid carcinoma; UCS, uterine carcinosarcoma; UVM, uveal melanoma; WT, wildtype

# Declarations

**Ethics approval and consent to participate**

Not applicable.

**Consent for publication**

Not applicable.

**Availability of data and material**

The dataset supporting the conclusions of this article is included within the article.

**Competing interests**

The authors declare that they have no competing interests.

**Funding**

This research and this article's publication costs were supported partially by the NCI Cancer Center Shared Resources (NIH-NCI P30CA54174 to YC), NIH (CTSA 1UL1RR025767-01 to YC, and R01GM113245 to YH), CPRIT (RP160732 to YC), and San Antonio Life Science Institute (SALSI Innovation Challenge Award 2016 to YH and YC). The funding sources had no role in the design of the study and collection, analysis, and interpretation of data and in writing the manuscript.

**Authors' contributions**

YCC, HHC, TZ, SZ, AG, LJW, YH, and YC conceived the study. YCC, YH, and YC designed the model. YCC performed data analysis. YCC, YH, and YC interpreted the data. YCC, HHC, TZ, SZ, AG, LJW, YH, and YC wrote and approved the final version of paper.

**Acknowledgements**

None.

# References

1. Hanahan D, Weinberg RA: **Hallmarks of cancer: the next generation**. *Cell* 2011, **144**(5):646-674.
2. Schmitt MW, Loeb LA, Salk JJ: **The influence of subclonal resistance mutations on targeted cancer therapy**. *Nat Rev Clin Oncol* 2016, **13**(6):335-347.
3. Phillips KA, Veenstra DL, Oren E, Lee JK, Sadee W: **Potential role of pharmacogenomics in reducing adverse drug reactions: a systematic review**. *JAMA : the journal of the American Medical Association* 2001, **286**(18):2270-2279.
4. Hertz DL, Rae J: **Pharmacogenetics of cancer drugs**. *Annu Rev Med* 2015, **66**:65-81.
5. Mina M, Raynaud F, Tavernari D, Battistello E, Sungalee S, Saghafinia S, Laessle T, Sanchez-Vega F, Schultz N, Oricchio E *et al*: **Conditional Selection of Genomic Alterations Dictates Cancer Evolution and Oncogenic Dependencies**. *Cancer cell* 2017, **32**(2):155-168 e156.
6. Park S, Lehner B: **Cancer type-dependent genetic interactions between cancer driver alterations indicate plasticity of epistasis across cell types**. *Molecular systems biology* 2015, **11**(7):824.
7. Iorio F, Knijnenburg TA, Vis DJ, Bignell GR, Menden MP, Schubert M, Aben N, Goncalves E, Barthorpe S, Lightfoot H *et al*: **A Landscape of Pharmacogenomic Interactions in Cancer**. *Cell* 2016, **166**(3):740-754.
8. Yang W, Soares J, Greninger P, Edelman EJ, Lightfoot H, Forbes S, Bindal N, Beare D, Smith JA, Thompson IR *et al*: **Genomics of Drug Sensitivity in Cancer (GDSC): a resource for therapeutic biomarker discovery in cancer cells**. *Nucleic Acids Res* 2013, **41**(Database issue):D955-961.
9. Alipanahi B, Delong A, Weirauch MT, Frey BJ: **Predicting the sequence specificities of DNA- and RNA-binding proteins by deep learning**. *Nat Biotechnol* 2015, **33**(8):831-838.
10. Zhou J, Troyanskaya OG: **Predicting effects of noncoding variants with deep learning-based sequence model**. *Nature methods* 2015, **12**(10):931-934.
11. Lin C, Jain S, Kim H, Bar-Joseph Z: **Using neural networks for reducing the dimensions of single-cell RNA-Seq data**. *Nucleic Acids Res* 2017, **45**(17):e156.
12. Angermueller C, Lee HJ, Reik W, Stegle O: **DeepCpG: accurate prediction of single-cell DNA methylation states using deep learning**. *Genome biology* 2017, **18**(1):67.
13. Chaudhary K, Poirion OB, Lu L, Garmire LX: **Deep Learning-Based Multi-Omics Integration Robustly Predicts Survival in Liver Cancer**. *Clinical cancer research : an official journal of the American Association for Cancer Research* 2017.
14. Wen M, Zhang Z, Niu S, Sha H, Yang R, Yun Y, Lu H: **Deep-Learning-Based Drug-Target Interaction Prediction**. *Journal of proteome research* 2017, **16**(4):1401-1409.
15. Kadurin A, Aliper A, Kazennov A, Mamoshina P, Vanhaelen Q, Khrabrov K, Zhavoronkov A: **The cornucopia of meaningful leads: Applying deep **adversarial autoencoders for new molecule development in oncology**. *Oncotarget* 2017, **8**(7):10883-10890.
16. Preuer K, Lewis RPI, Hochreiter S, Bender A, Bulusu KC, Klambauer G: **DeepSynergy: Predicting anti-cancer drug synergy with Deep Learning**. *Bioinformatics* 2017.
17. Patro R, Duggal G, Love MI, Irizarry RA, Kingsford C: **Salmon provides fast and bias-aware quantification of transcript expression**. *Nature methods* 2017, **14**(4):417-419.
18. Newton Y, Novak AM, Swatloski T, McColl DC, Chopra S, Graim K, Weinstein AS, Baertsch R, Salama SR, Ellrott K *et al*: **TumorMap: Exploring the Molecular Similarities of Cancer Samples in an Interactive Portal**. *Cancer Res* 2017, **77**(21):e111-e114.
19. Barretina J, Caponigro G, Stransky N, Venkatesan K, Margolin AA, Kim S, Wilson CJ, Lehar J, Kryukov GV, Sonkin D *et al*: **The Cancer Cell Line Encyclopedia enables predictive modelling of anticancer drug sensitivity**. *Nature* 2012, **483**(7391):603-607.
20. Cancer Cell Line Encyclopedia C, Genomics of Drug Sensitivity in Cancer C: **Pharmacogenomic agreement between two cancer cell line data sets**. *Nature* 2015, **528**(7580):84-87.
21. Kowarik A, Templ M: **Imputation with the R Package VIM**. *2016* 2016, **74**(7):16.
22. Alfons A, Templ M: **Estimation of Social Exclusion Indicators from Complex Surveys: The R Package laeken**. *2013* 2013, **54**(15):25.
23. He K, Zhang X, Ren S, Sun J: **Delving deep into rectifiers: Surpassing human-level performance on imagenet classification**. In: *Proceedings of the IEEE international conference on computer vision: 2015*; 2015: 1026-1034.
24. Pumperla M: **Keras + Hyperopt: A very simple wrapper for convenient hyperparameter optimization**. In.; 2016.
25. Shetzer Y, Solomon H, Koifman G, Molchadsky A, Horesh S, Rotter V: **The paradigm of mutant p53-expressing cancer stem cells and drug resistance**. *Carcinogenesis* 2014, **35**(6):1196-1208.
26. Hientz K, Mohr A, Bhakta-Guha D, Efferth T: **The role of p53 in cancer drug resistance and targeted chemotherapy**. *Oncotarget* 2017, **8**(5):8921-8946.
27. Wang JB, Dong DF, Wang MD, Gao K: **IDH1 overexpression induced chemotherapy resistance and IDH1 mutation enhanced chemotherapy sensitivity in Glioma cells in vitro and in vivo**. *Asian Pac J Cancer Prev* 2014, **15**(1):427-432.
28. Bywater MJ, Poortinga G, Sanij E, Hein N, Peck A, Cullinane C, Wall M, Cluse L, Drygin D, Anderes K *et al*: **Inhibition of RNA polymerase I as a therapeutic strategy to promote cancer-specific activation of p53**. *Cancer Cell* 2012, **22**(1):51-65.
29. Hein N, Cameron DP, Hannan KM, Nguyen NN, Fong CY, Sornkom J, Wall M, Pavy M, Cullinane C, Diesch J *et al*: **Inhibition of Pol I transcription treats murine and human AML by targeting the leukemia-initiating cell population**. *Blood* 2017, **129**(21):2882-2895.

30. Wei JC, Meng FD, Qu K, Wang ZX, Wu QF, Zhang LQ, Pang Q, Liu C: **Sorafenib inhibits proliferation and invasion of human hepatocellular carcinoma cells via up-regulation of p53 and suppressing FoxM1**. *Acta Pharmacol Sin* 2015, **36**(2):241-251.
31. Ling X, Calinski D, Chanan-Khan AA, Zhou M, Li F: **Cancer cell sensitivity to bortezomib is associated with survivin expression and p53 status but not cancer cell types**. *J Exp Clin Cancer Res* 2010, **29**:8.
32. Li P, Zhao M, Parris AB, Feng X, Yang X: **p53 is required for metformin-induced growth inhibition, senescence and apoptosis in breast cancer cells**. *Biochem Biophys Res Commun* 2015, **464**(4):1267-1274.
33. Huang da W, Sherman BT, Lempicki RA: **Systematic and integrative analysis of large gene lists using DAVID bioinformatics resources**. *Nature protocols* 2009, **4**(1):44-57.
34. Huang da W, Sherman BT, Lempicki RA: **Bioinformatics enrichment tools: paths toward the comprehensive functional analysis of large gene lists**. *Nucleic acids research* 2009, **37**(1):1-13.
35. Fulton B, Spencer CM: **Docetaxel. A review of its pharmacodynamic and pharmacokinetic properties and therapeutic efficacy in the management of metastatic breast cancer**. *Drugs* 1996, **51**(6):1075-1092.
36. Drygin D, Lin A, Bliesath J, Ho CB, O'Brien SE, Proffitt C, Omori M, Haddach M, Schwaebe MK, Siddiqui-Jain A *et al*: **Targeting RNA polymerase I with an oral small molecule CX-5461 inhibits ribosomal RNA synthesis and solid tumor growth**. *Cancer research* 2011, **71**(4):1418-1430.
37. Xu H, Di Antonio M, McKinney S, Mathew V, Ho B, O'Neil NJ, Santos ND, Silvester J, Wei V, Garcia J *et al*: **CX-5461 is a DNA G-quadruplex stabilizer with selective lethality in BRCA1/2 deficient tumours**. *Nature communications* 2017, **8**:14432.
38. Angermueller C, Parnamaa T, Parts L, Stegle O: **Deep learning for computational biology**. *Molecular systems biology* 2016, **12**(7):878.
39. Yosinski J, Clune J, Nguyen A, Fuchs T, Lipson H: **Understanding neural networks through deep visualization**. *arXiv preprint arXiv:150606579* 2015.
40. Shrikumar A, Greenside P, Kundaje A: **Learning important features through propagating activation differences**. *arXiv preprint arXiv:170402685* 2017.
41. Kalinin AA, Higgins GA, Reamaroon N, Soroushmehr S, Allyn-Feuer A, Dinov ID, Najarian K, Athey BD: **Deep Learning in Pharmacogenomics: From Gene Regulation to Patient Stratification**. *arXiv preprint arXiv:180108570* 2018.

# Tables

**Table 1 - Performance of our DNNs and other models**

| Measurement | Model | Linear regression | SVM | Random initialization | PCA | $E_{enc}$ only | $M_{enc}$ only |
|---|---|---|---|---|---|---|---|
| Median MSE in testing samples[a] | 1.96 | 10.24[b] | 8.92[c] | 2.30 | 2.44 | 1.96 | 3.09 |
| Median number of training epochs[a] | 14 | -- | -- | 9 | 29 | 17 | 9.5 |

[a]Median of 100 shuffles of training, validation, and testing samples
[b]Result of one multivariate regression model
[c]Results of 265 SVM models, each predicting $IC_{50}$ for a drug

**Table 2 - Top mutations in modulating drug response among individual cancers**

| Cancer | Gene | Mutation rate | Num. modulated drugs | Num. sensitive drugs | Num. resistant drugs |
|---|---|---|---|---|---|
| **LUAD** | *TP53* | 46.1% | 235 | 0 | 235 |
| **LUSC** | *TP53* | 75.1% | 228 | 0 | 228 |
| **STAD** | *TP53* | 43.3% | 224 | 0 | 224 |
| **HNSC** | *TP53* | 66.1% | 207 | 0 | 207 |
| **COAD** | *TP53* | 55.7% | 197 | 0 | 197 |
| **LIHC** | *TP53* | 27.0% | 194 | 1 | 193 |
| **BRCA** | *TP53* | 32.2% | 182 | 7 | 175 |
| **LGG** | *IDH1* | 77.3% | 159 | 138 | 21 |
| **PRAD** | *TP53* | 10.8% | 146 | 1 | 145 |
| **KIRC** | *PBRM1* | 38.0% | 142 | 3 | 139 |

**Table 3 - Top gene mutations modulating pan-cancer drug response**

| Gene | Mutation rate | Num. modulated drugs | Num. sensitive drugs | Num. resistant drugs |
|---|---|---|---|---|
| *TP53* | 34.3% | 251 | 9 | 242 |
| *CSMD3* | 12.6% | 223 | 12 | 211 |
| *SYNE1* | 11.5% | 218 | 10 | 208 |
| *TTN* | 30.2% | 206 | 44 | 162 |
| *RYR2* | 11.9% | 199 | 14 | 185 |
| *USH2A* | 10.7% | 191 | 12 | 179 |
| *LRP1B* | 12.1% | 188 | 19 | 169 |
| *FLG* | 11.0% | 183 | 9 | 174 |
| *MUC16* | 19.5% | 161 | 51 | 110 |
| *PCLO* | 10.5% | 155 | 12 | 143 |
| *PIK3CA* | 11.7% | 144 | 45 | 99 |

**Table 4 - Top GO clusters enriched in top 300 differentially expressed genes associated with predicted response to docetaxel**

| GO ID | GO term | Num. genes | *P*-value |
|---|---|---|---|
| **Cluster 1 (enrichment score: 10.89)** | | | |
| GO:0007049 | cell cycle | 40 | $1.13 \times 10^{-10}$ |
| GO:0022402 | cell cycle process | 33 | $3.51 \times 10^{-10}$ |
| GO:0000279 | M phase | 32 | $1.01 \times 10^{-15}$ |
| **Cluster 2 (enrichment score: 3.96)** | | | |
| GO:0000166 | nucleotide binding | 56 | $1.95 \times 10^{-4}$ |
| GO:0032553 | ribonucleotide binding | 54 | $2.74 \times 10^{-6}$ |
| GO:0032555 | purine ribonucleotide binding | 54 | $2.74 \times 10^{-6}$ |
| **Cluster 3 (enrichment score: 3.45)** | | | |
| GO:0000278 | mitotic cell cycle | 26 | $1.01 \times 10^{-9}$ |
| GO:0051726 | regulation of cell cycle | 15 | $8.48 \times 10^{-4}$ |
| GO:0007346 | regulation of mitotic cell cycle | 12 | $3.09 \times 10^{-5}$ |
| **Cluster 4 (enrichment score: 2.47)** | | | |
| GO:0051327 | M phase of meiotic cell cycle | 8 | $9.46 \times 10^{-4}$ |
| GO:0007126 | meiosis | 8 | $9.46 \times 10^{-4}$ |
| GO:0051321 | meiotic cell cycle | 8 | $1.07 \times 10^{-3}$ |
| **Cluster 5 (enrichment score: 2.07)** | | | |
| GO:0051276 | chromosome organization | 13 | $8.64 \times 10^{-2}$ |
| GO:0007059 | chromosome segregation | 6 | $9.34 \times 10^{-3}$ |
| GO:0000070 | mitotic sister chromatid segregation | 5 | $2.45 \times 10^{-3}$ |

Each cluster is represented by the largest three GO terms.

**Table 5 - Top GO clusters enriched in top 300 differentially expressed genes associated with predicted response to CX-5461**

| GO ID | GO term | Num. genes | *P*-value |
|---|---|---|---|
| **Cluster 1 (enrichment score: 8.65)** | | | |
| GO:0043062 | extracellular structure organization | 17 | $2.93 \times 10^{-9}$ |
| GO:0030198 | extracellular matrix organization | 15 | $4.55 \times 10^{-10}$ |
| GO:0005201 | extracellular matrix structural constituent | 13 | $2.64 \times 10^{-9}$ |
| **Cluster 2 (enrichment score: 6.13)** | | | |
| GO:0008544 | epidermis development | 18 | $2.35 \times 10^{-9}$ |
| GO:0007398 | ectoderm development | 18 | $7.71 \times 10^{-9}$ |
| GO:0030855 | epithelial cell differentiation | 8 | $4.60 \times 10^{-3}$ |
| **Cluster 3 (enrichment score: 4.23)** | | | |
| GO:0030199 | collagen fibril organization | 9 | $7.34 \times 10^{-9}$ |
| GO:0032963 | collagen metabolic process | 6 | $5.37 \times 10^{-5}$ |
| GO:0044259 | multicellular organismal macromolecule metabolic process | 6 | $8.96 \times 10^{-5}$ |
| **Cluster 4 (enrichment score: 2.84)** | | | |
| GO:0006928 | cell motion | 18 | $8.22 \times 10^{-4}$ |
| GO:0016477 | cell migration | 13 | $9.51 \times 10^{-4}$ |
| GO:0048870 | cell motility | 13 | $2.33 \times 10^{-3}$ |
| **Cluster 5 (enrichment score: 2.60)** | | | |
| GO:0060429 | epithelium development | 12 | $6.39 \times 10^{-4}$ |
| GO:0030855 | epithelial cell differentiation | 8 | $4.60 \times 10^{-3}$ |
| GO:0009913 | epidermal cell differentiation | 6 | $4.49 \times 10^{-3}$ |

Each cluster is represented by the largest three GO terms.

# Figures

**Figure 1 - Illustration of the proposed neural network model**

(A) Model overview. Mutation and expression data of TCGA (n = 9,059) were used to pre-train two autoencoders (highlighted in blue and green) to extract data representations. Encoders of the autoencoders, namely mutation encoder $M_{enc}$ and expression encoder $E_{enc}$, were linked to a prediction network (P; denoted in orange) and the entire model (*i.e.*, $M_{enc}$, $E_{enc}$, and P) was trained using CCLE data (n = 622, of which 80%, 10%, and 10% used as training, validation, and testing, respectively) to predict the response to 265 drugs. (B) Architecture of the neural networks. Numbers denote the number of neurons at each layer.

**Figure 2 - Model construction and evaluation using CCLE datasets**

(A) Density plots of true (with missing values), imputed, and predicted $IC_{50}$ data of CCLE and predicted data of TCGA. (B) Heatmaps of imputed and predicted $IC_{50}$ data of CCLE. (C, D) Sample-wise Pearson and Spearman correlation between imputed and predicted $IC_{50}$ data of CCLE samples. (E) Mean square errors of our and 4 other DNN-based designs. The proposed model was compared to a model with no TCGA pre-training (with encoders randomly initialized; abbreviated as Rand Init), with encoders substituted by PCAs, with $E_{enc}$ only (no $M_{enc}$), and with $M_{enc}$ only (no $E_{enc}$). Each model was trained for 100 times, each of which CCLE samples were randomly assigned into training, validation, and testing sets.

**Figure 3 - Associations of gene mutations to predicted drug response in TCGA – per-cancer study**

(A) Predicted $IC_{50}$ of TCGA tumors with known drug targets in a cancer type. Significance of $\Delta IC_{50}$ between tumors with and without a gene mutation was assessed by the two-tailed *t*-test. (B) Gene mutations significantly associated with predicted drug response in a cancer type. Middle panel, significant mutation–drug pairs in each cancer (with Bonferroni adjusted *t*-test $P < 1.0\times10^{-5}$). Nodes labeled with names are those with extreme significance (adjust $P < 1.0\times10^{-60}$) and magnitude of $\Delta IC_{50}$ ($|\Delta IC_{50}| \geq 0.5$). Top 10 cancer types with the largest sample sizes are denoted by node color and shape. (C) Box plots of three mutation–drug examples in BRCA and LGG.

**Figure 4 - Associations of gene mutations to predicted drug response in TCGA – pan-cancer study**

(A) Gene mutations significantly associated with predicted drug response across all TCGA samples. Here only the 11 genes with mutation rates larger than 10% were analyzed. Nodes labeled with names are those with extreme significance (adjust $P < 1.0\times10^{-200}$) and magnitude of $\Delta IC_{50}$ ($\Delta IC_{50} \geq 0.7$ or $\Delta IC_{50} < 0$). (B, C) Examples of drugs modulated by *TP53* and *TTN* mutations, respectively.

**Figure 5 - Pharmacogenomics analysis of docetaxel and CX-5461 in TCGA**

(A) Waterfall plot of predicted $IC_{50}$ for the two drugs across all TCGA samples. Tumors with extreme $IC_{50}$ values (top and bottom 1%) were denoted as the resistant and sensitive groups. (B) Cancer type composition of resistant and sensitive samples. Cancer types accounted for at least 10% in any group are highlighted in bold and shown in (C). (C) Heatmaps of cancer type composition, top differentially mutated genes, and top differentially expressed genes between the two groups. In the expression heatmap, genes are normalized and hierarchically clustered, and samples are clustered within each group.

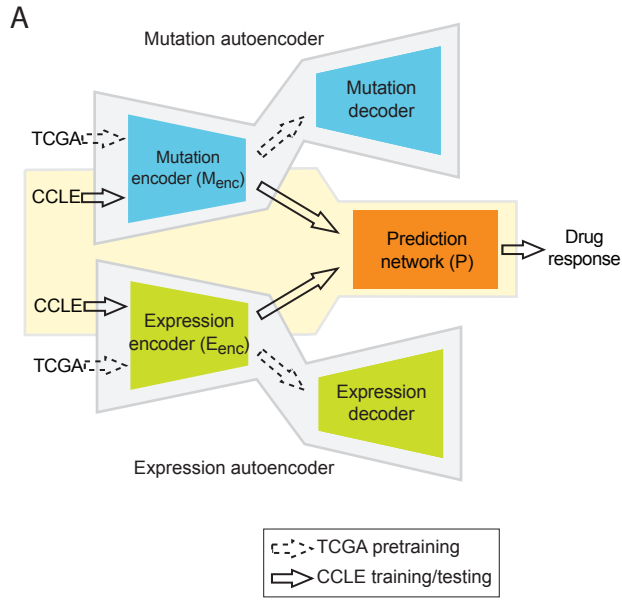
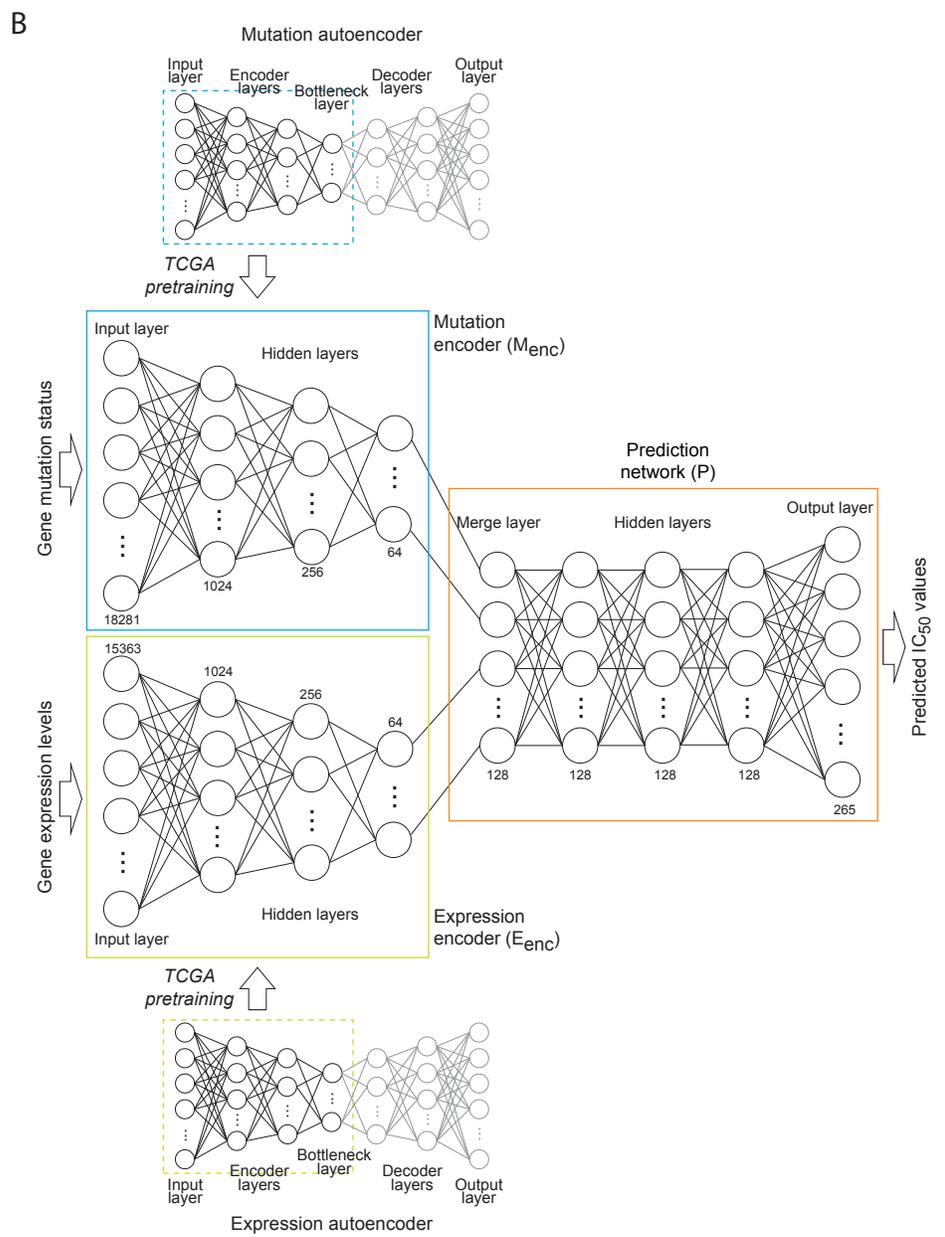

Figure 1

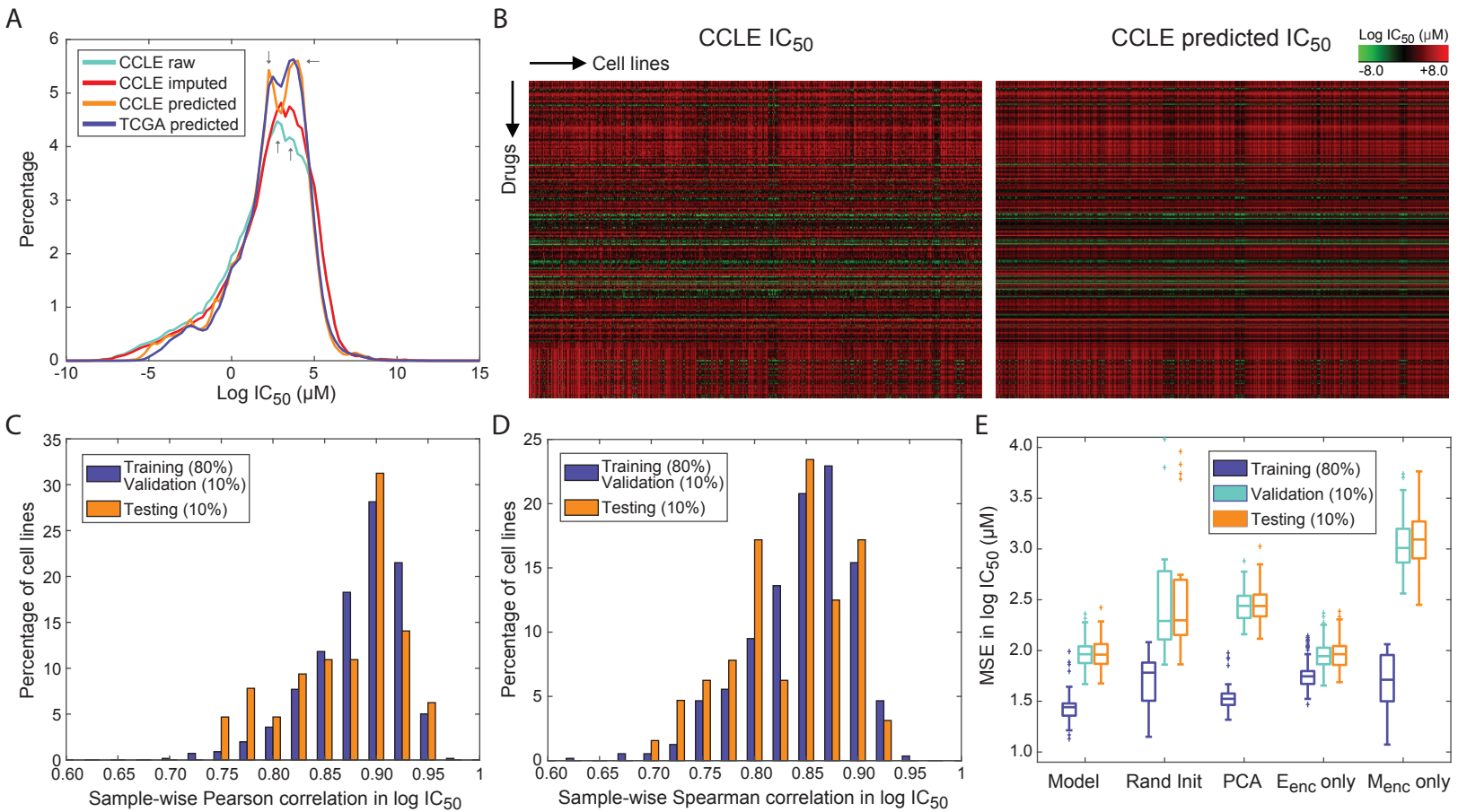

Figure 2

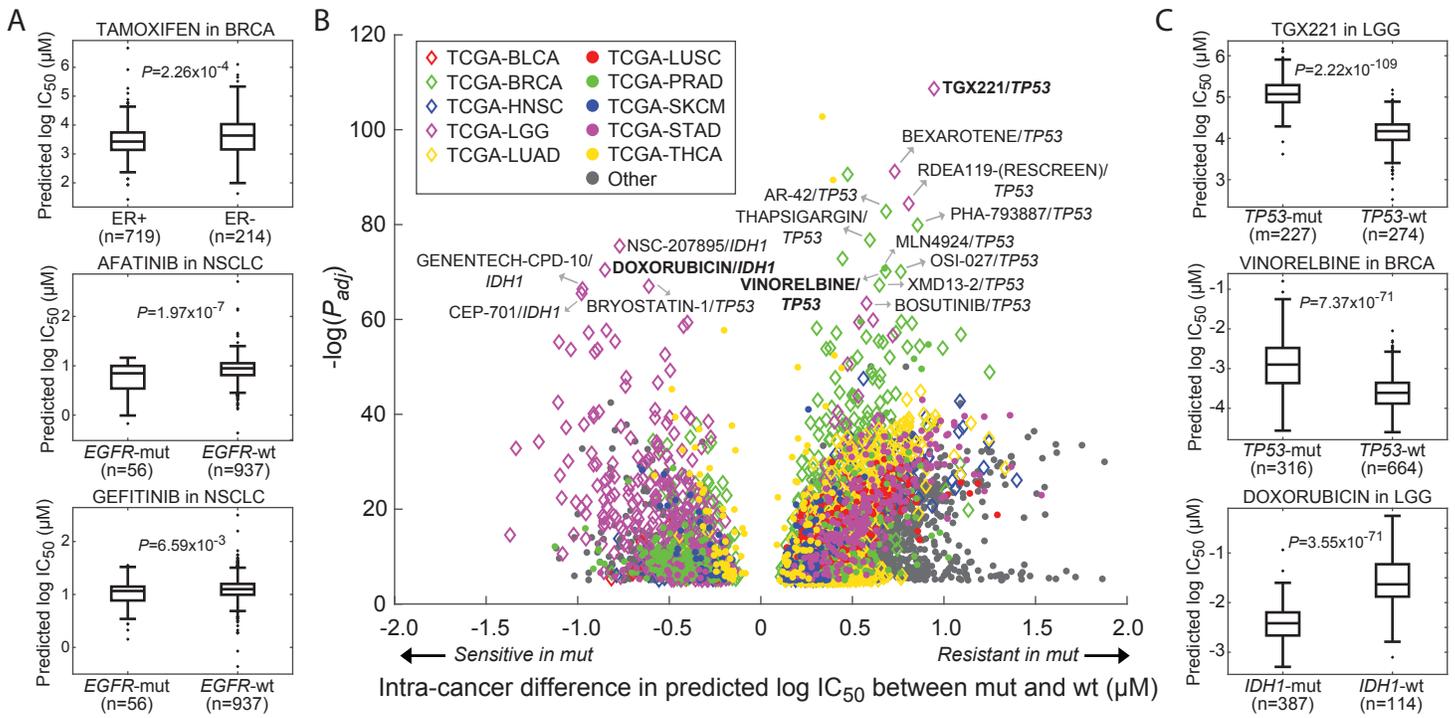

Figure 3

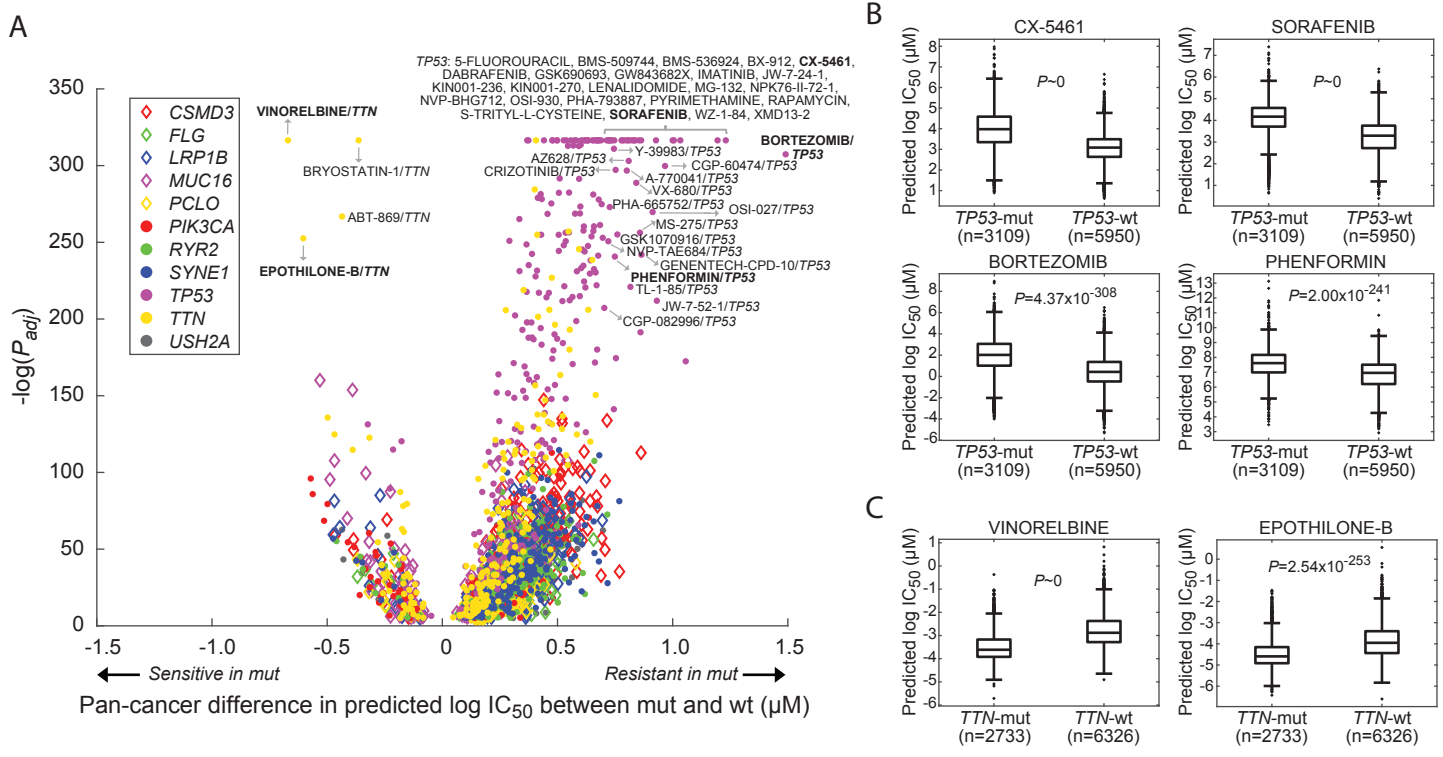

Figure 4

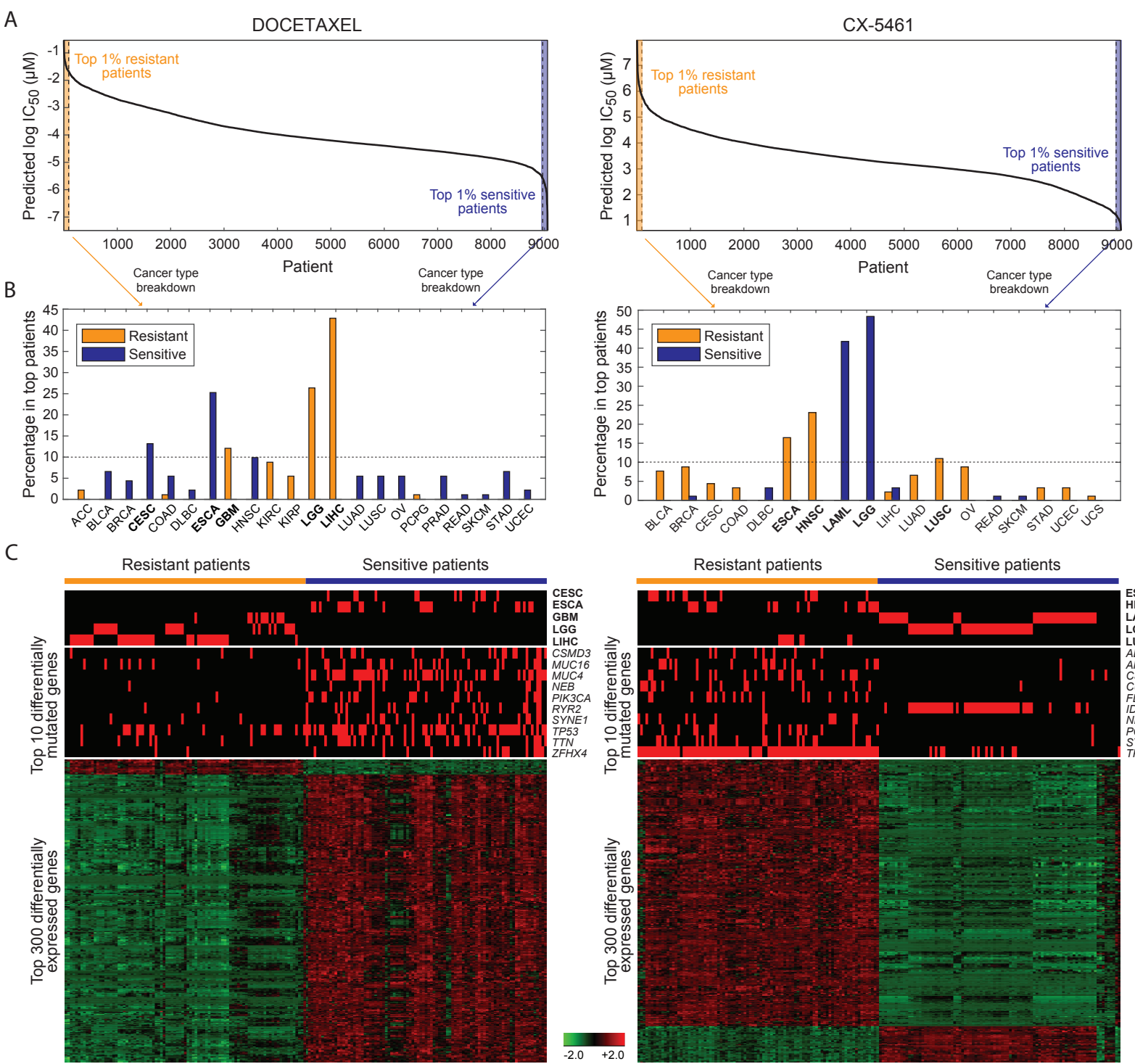

Figure 5